\pdfoutput=1

\documentclass{article}
\usepackage{stfloats}  
\usepackage{graphicx}

\usepackage[final]{neurips_2023}

\usepackage[utf8]{inputenc} 
\usepackage[T1]{fontenc}    
\usepackage{hyperref}       
\usepackage{url}            
\usepackage{booktabs}       
\usepackage{amsfonts}       
\usepackage{nicefrac}       
\usepackage{microtype}      
\usepackage{xcolor}         
\usepackage{lmodern}

\usepackage{listings}
\usepackage{tcolorbox}
\usepackage{amsmath}
\usepackage{enumitem}

\usepackage{tabularx} 
\usepackage{colortbl} 
\usepackage{booktabs} 

\tcbuselibrary{skins}
\tcbuselibrary{raster}

\definecolor{codegreen}{rgb}{0,0.6,0}
\definecolor{codegray}{rgb}{0.5,0.5,0.5}
\definecolor{codepurple}{rgb}{0.58,0,0.82}
\definecolor{backcolour}{rgb}{0.95,0.95,0.92}
\definecolor{codeblue}{rgb}{0.13,0.13,1}
\definecolor{codered}{rgb}{0.2,0.2,0.2.08}
\definecolor{codedark}{rgb}{0.85,0.85,0.85}
\lstdefinestyle{mystyle}{
    backgroundcolor=\color{backcolour},   
    commentstyle=\color{codegreen},
    keywordstyle=\color{codeblue},
    numberstyle=\tiny\color{codegray},
    stringstyle=\color{codepurple},
    basicstyle=\ttfamily\footnotesize,
    breakatwhitespace=false,         
    breaklines=true,                 
    captionpos=b,                    
    keepspaces=true,              
    numbersep=5pt,                  
    showspaces=false,                
    showstringspaces=false,
    showtabs=false,                  
    tabsize=2,
    frame=single,                    
    rulecolor=\color{codered},       
    escapeinside={\%*}{*)}           
}

\lstdefinestyle{responsestyle}{
    backgroundcolor=\color{codedark},   
    commentstyle=\color{codegreen},
    keywordstyle=\color{codeblue},
    numberstyle=\tiny\color{codegray},
    stringstyle=\color{codepurple},
    basicstyle=\ttfamily\footnotesize,
    breakatwhitespace=false,         
    breaklines=true,                 
    captionpos=b,                    
    keepspaces=true,                    
    numbersep=5pt,                  
    showspaces=false,                
    showstringspaces=false,
    showtabs=false,                  
    tabsize=2,
    frame=none,                    
    rulecolor=\color{codered},       
    escapeinside={\%*}{*)}           
}

\definecolor{codecolor}{rgb}{0.9,0.1,0.1} 

\definecolor{lightgray}{gray}{0.9}

\setcitestyle{authoryear,open={(},close={)}} 

\renewcommand{\cite}{\citep}

\title{DeepDecipher: Accessing and Investigating Neuron Activation in Large Language Models}

\author{%
  Albert Garde$^{*}$\\
  Apart Research\\
  \text{albertsgarde@gmail.com}\\
  \And
  Esben Kran$^{*}$\\
  Apart Research\\
\text{esben@apartreserach.com}
  \And
  Fazl Barez\\
  Apart Research\\
  Edinburgh Centre for Robotics \\
  Department of Engineering Sciences, University of Oxford\\
  \text{fazl@robots.ox.ac.uk}
}

\begin{document}
\maketitle
\def\thefootnote{*}\footnotetext{Equal contribution.}

\begin{abstract}

As large language models (LLMs) become more capable, there is an urgent need for interpretable and transparent tools. Current methods are difficult to implement, and accessible tools to analyze model internals are lacking.
To bridge this gap, we present DeepDecipher - an API and interface for probing neurons in transformer models' MLP layers. DeepDecipher makes the outputs of advanced interpretability techniques for LLMs readily available. The easy-to-use interface also makes inspecting these complex models more intuitive.
This paper outlines DeepDecipher's design and capabilities. We demonstrate how to analyze neurons, compare models, and gain insights into model behavior. For example, we contrast DeepDecipher's functionality with similar tools like Neuroscope and OpenAI's Neuron Explainer.
DeepDecipher enables efficient, scalable analysis of LLMs. By granting access to state-of-the-art interpretability methods, DeepDecipher makes LLMs more transparent, trustworthy, and safe. Researchers, engineers, and developers can quickly diagnose issues, audit systems, and advance the field.

\end{abstract}

\section{Introduction}
Explainability and safety in AI are becoming increasingly important as language models like GPT-4 make up the next generation of software \cite{baktash2023gpt4, Wu2023BloombergGPTAL}. Methods in explainable AI, making AI more understandable for human users, often draw inspiration from interpretability research. Recent work has shifted attention from vision models \cite{erhan_visualizing_2009, simonyan_deep_2014} to Transformer and generative models \cite{vaswani2023attention, elhage2021mathematical}.

Transformers contain attention and multi-layer perceptron (MLP) layers \cite{Brown2020LanguageMA}. The attention layers (that "transfer information" between sections of the text in latent space) have received more attention within the interpretability research community \cite{olsson2022context}, while the MLP layers (that non-linearly modify this information) remain underexplored \cite{elhage2022solu}. Given their role in complex language tasks, the lack of targeted interpretability tools for MLPs is a glaring gap in ensuring explainable AI becomes the norm.
\begin{figure}[ht]
\includegraphics[width=1\linewidth]{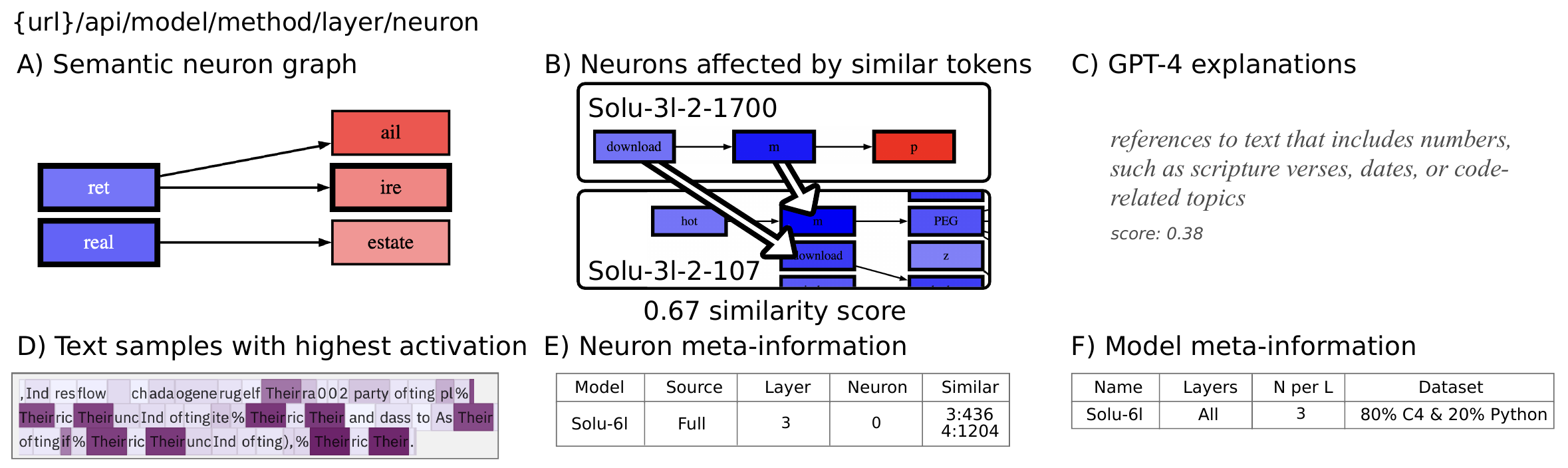}
\caption{
Data sources available in the DeepDecipher user interface and API: \textbf{A)} Graphs encapsulating which tokens influence neuron activation \cite{foote_neuron_2023}.
\textbf{B)} The graphs from (A) are compared, and co-occurrence of token nodes is used to calculate the similarity score as the two-way maximum proportion of overlapping nodes.
\textbf{C)} GPT-4-based automated descriptions of the semantics associated with neuron activation \cite{bills_language_2023}.
\textbf{D)} A list of snippets that this neuron activates highly to. Each snippet consists of 1024 tokens colored according to the neuron activation on that token \cite{nanda_neuroscope_2022}.
\textbf{E)} Statistics and meta-information about the neuron
\textbf{F)} Meta-information about the model and layers
}
\label{fig:viz_overview}
\end{figure}
In this paper, we introduce and describe DeepDecipher, a tool to visualize, interpret, and explain MLP neuron activation in large language models. It eases access to explanations of low-level causal components of open language models \cite{gpt2, pythia} based on the principles of mechanistic interpretability \cite{olah2020zoom}. DeepDecipher does not introduce novel methods for neuron explanations but makes recent interpretability research results available in an accessible user interface and API.

\section{Functionality}
DeepDecipher aims to simplify access to the data generated by existing interpretability methods, specifically methods that provide information on the behaviour of MLP neurons in transformer-based language models. It does this both through an API and a visual interface that allows users to understand when and why an MLP neuron activates. DeepDecipher provides access to the following modern MLP interpretability methods.

\textbf{Neuron to Graph (N2G)}  is a method to create a graph of tokens sequences that affect a specific MLP neuron's activation. It is generated by backtracking from the most activating tokens and finding the tokens that affect the activation on the final token the most by taking examples of sequences that lead to activation and replacing tokens until one finds 1) the tokens most important for activation on 2) the end tokens where the neuron activates.

\textbf{Neuroscope} is a database of the 20 1024 token sequences that each neuron of a model activates the most to. The 1024 token length sequences are sampled from OpenWebText \cite{Gokaslan2019OpenWeb}, the Pile \cite{pile}, C4 and Python depending on the training dataset of the models. The model activations on runs over the token sequences are used to generate this dataset: For OpenWebText, each model is run over 9b tokens; for the Pile, each model is run over 2b tokens and for C4 and Python, each model is run over 1.4b tokens of C4 and 0.3b tokens of Python.

\textbf{Neuron Explainer} is an automated interpretability method that uses GPT-4 to explain which categories of token sequences a neuron responds to (e.g. "references to movies, characters, and entertainment.") and evaluates how well GPT-4 can predict that neuron's activation based on this description. They apply their method to the open GPT-2 XL model.

Based on the research behind N2G, we also introduce a \textbf{search function} to find the neurons that respond the most to a particular token along with a \textbf{neuron similarity} based on the same N2G data. The search queries the N2G graphs for the token and similarity is scored based on co-occurring tokens in the N2G graphs of two neurons.

\textbf{Website:}
The website provides an interactive visual interface to enable rapid hypothesis testing and model understanding. As shown in Figure \ref{fig:viz_overview}, visual representations are provided for all model data accessible through the API. This allows users to visually explore the inner workings of the model.

Specifically, it is possible to navigate through the model by neuron index and search for neurons that activate on specific tokens or concepts. The visualizations make it easy to identify patterns, trends, and relationships in the model's representations and processing. Users can quickly validate assumptions and gain insights by observing model behavior.
The interface provides access to different types of data about the model, as outlined in Table \ref{tab:available_data}. With these visualization and data analysis tools, researchers can interactively probe the model to test hypotheses about its functioning. The hands-on exploration facilitates rapid intuition development about model mechanisms.

\textbf{API:}
The data found on the website is all available in JSON format through the API.
Endpoints take the form \lstinline|/api/<model>/<service>/<layer>/<neuron>| where \lstinline|<service>| is the type of data e.g. \lstinline|neuron2graph|, \lstinline|neuroscope| or \lstinline|all|.

For examples, see Section \ref{sec:api_usage} or the repository\footnote{\href{https://github.com/apartresearch/deepdecipher}{https://github.com/apartresearch/deepdecipher}}.
\textbf{Extendable framework:}
Everything is designed to be easily extendable with additional models and interpretability methods. The server is designed with scalability in mind to handle many simultaneous users.

\begin{table}[ht]
    \centering
    \begin{tabularx}{\textwidth}{c|X|X|X}
        \toprule[1.5pt]
        \rowcolor{lightgray}
        & \textbf{DeepDecipher} & \textbf{Neuroscope} & \textbf{Neuron Explainer} \\
        \midrule[1pt]
        \textbf{Unique Contribution} & Search and availability & Activating dataset examples & GPT-4 neuron explanations \\
        \textbf{API Output} & JSON & XML (HTML) & JSON \\
        \textbf{Speed (requests/second)} & 56 (API)* & 87 (web page) & 53 (API) \\
        \textbf{Data Available} & Neuron to Graph, Neuroscope, Model Explanations (extendable) & Neuroscope & Model Explanations \\
        \textbf{Models Available} & 25 (extendable) & 25 & 2 \\
        \bottomrule[1.5pt]
    \end{tabularx}
    \caption{Comparison of DeepDecipher, Neuroscope, and OpenAI Neuron Explanations. Note that all measurements are mainly bottle-necked by client download speed and do therefore not represent the capacity of the server. *The stated API speed is the lowest speed across services. See table \ref{tab:available_data} for details.}
    \label{tab:speed}
\end{table}


\section{Available data}\label{sec:available_data}
We currently support 3 services and 25 models.
The models supported are the same as those supported by Neuroscope \cite{nanda_neuroscope_2022}, which include a range of small \texttt{solu} and \texttt{gelu} models \cite{nanda_progress_2023}, 4 GPT-2 models \cite{gpt2}, and 3 \texttt{pythia} models \cite{pythia}.
Full details are available on the website\footnote{\url{https://deepdecipher.org/viz}}.
Neuroscope is supported for all models, Neuron2Graph is on its way for all models, and Neuron Explainer is only available for GPT-2 Small and GPT-2 XL.

\begin{table}[ht]
    \centering
    \begin{tabular}{p{0.19\textwidth}|p{0.46\textwidth}|p{0.16\textwidth}|p{0.06\textwidth}}
        \toprule[1.5pt]
        \rowcolor{lightgray}
        & \textbf{Description} & \textbf{Models} & \textbf{API Speed} \\
        \midrule[1pt]
        \textbf{Neuron to Graph} & Graph illustrating what tokens activate or modulate a neuron \cite{foote_neuron_2023} & \verb|solu-6l-pile| (all coming*) & 388 \\
        \textbf{Neuroscope} & 20 top-activating 1024-token sequences based on the training data \cite{nanda_neuroscope_2022} & All & 56\\
        \textbf{Neuron Explainer} & Explanation by GPT-4 and score of the explanation \cite{bills_language_2023} & \verb|gpt2-small| and \verb|gpt2-xl| & 384 \\
        \bottomrule[1.5pt]
    \end{tabular}
    \caption{Data available in the model. API speed is measured as requests per second. Note that all measurements are mainly bottle-necked by client download speed and do therefore not represent the capacity of the server.
    *Neuron to Graph data is currently only available for a single model, but we plan to extend this to all models very soon.}
    \label{tab:available_data}
\end{table}

\section{Usage}
DeepDecipher allows users with a comprehensive range of capabilities, including:
\begin{enumerate}
\item DeepDecipher offers API access to state-of-the-art data concerning the functionality, activation patterns, and activation dependencies of individual MLP neurons in various open large language models. Additionally, it provides the flexibility to easily expand this API to accommodate new models.
\item With DeepDecipher, you can efficiently search for neurons that respond to specific tokens within these models.
\item DeepDecipher creates dedicated web pages for each model, layer, and MLP neuron, complete with static links that grant users convenient access to the API data. Furthermore, DeepDecipher can be deployed locally, making it compatible with proprietary models.
\end{enumerate}
DeepDecipher's functionality proves useful under many scenarios (listed below). For mechanistic interpretability researchers, the API provides access to useful data on MLP neuron functionality without the need to re-run the computationally expensive processing (e.g., $O(n^{2/3})$ for \cite{bills_language_2023}).
The upcoming EU AI Act \cite{EUAIAct} ushers in an era of increased transparency for "high-risk AI" models. When these models are deployed in sensitive applications, users will have the right to understand why certain outputs or decisions were made. As engineers prepare for this changing regulatory environment, tools like DeepDecipher can provide helpful capabilities to enable model interpretability.
Web pages, API, and search functionality provide the scaffolding to explain the output of neural networks \cite{MONTAVON20181, conmy_towards_2023}.
As an engineer and researcher, we can rapidly test theories about language model comprehension. Below, we present some case studies.
Using the search functionality for the \texttt{solu-6l-pile} model\footnote{Available on the website's \url{https://deepdecipher.org/viz/solu-6l-pile/all} solu-6l DeepDecipher page}:
\begin{itemize}
\item Seeing how Spanish is represented compared to English, finding that no tokens encode for \texttt{hola} (0) compared to \texttt{hello} (7) which corresponds to the training dataset being filtered for the English language \cite{pile}.
\item Evaluating the emotional specificity of a model proxied by how many neurons respond to happy (9), sad (5), hate (3), love (20), morose (0) and other emotions, corroborating benchmarks for emotional intelligence of large language models \cite{Wang2023-hw}.
\item Understanding how models differentiate between homographs, e.g., \texttt{Apple} the company and \texttt{apple} the fruit. By looking at the token activation dependency graphs on each page, we can see that of the 8 neurons responding to \texttt{apple}, 5 are about the company, 2 are about the fruit and 1 is ambiguous, showing disambiguation between the two concepts \cite{Mikolov2013-kt}.
\end{itemize}
For application developers using language models, providing easy access for users to embedded and linked explanations for the neurons that are involved in actions the AI system takes enables a more secure and explainable user-AI interface \cite{Vig2019-xc}.
See Appendix \ref{sec:api_usage} for documentation on using the API.

\section{Discussion and future work}
The DeepDecipher project exposes the internals of neural networks in an interface and aims to be a robust and reliable application for users interested in understanding what happens within LLMs. The vision for DeepDecipher is to be an integrated component to make any frontier AI system more explainable. This includes redesigning parts of the user interface, ensuring support for proprietary models, implementing better search functionality and partnering with researchers to provide a comprehensive database of models.

\section{Conclusions}
DeepDecipher provides a user-friendly way to understand how neurons in Transformer and MLP layers work. It aims to make complex AI systems easier to explain and more secure. Looking ahead, DeepDecipher is focusing on making it simpler to incorporate its features into other research projects and applications, becoming a mainstay of explainable applications.

\section*{Ackowledgements}
We are grateful to Alex Foote for his continued assistance in getting Neuron2Graph integrated well in DeepDecipher.
We thank Miko\l{}ai Kniejski for our long discussions on how to design the websites UX.
Bart Buss, for his patience answering our questions on what features would be useful.
DTU (Denmarks Technical University) for use of their HPC cluster for running the Neuron2Graph algorithm on all the models.

\bibliography{bib}
\bibliographystyle{acl_natbib} 

\appendix

\section{Limitations}
Mechanistic interpretability for practical usage in safety monitoring of large language models (LMMs) is limited in its applicability due to the large amount of neurons in state-of-the-art LLMs such as GPT-4 \cite{openai2023gpt4} and the difficulty in reverse-engineering the learned functions.
DeepDecipher does not directly address this issue, but since it provides easy access to data on many neurons, it can hopefully aid the process regardless.

All three methods rely on max activating examples.
The neurons' behaviour here may not be representative of its behaviour in general.
This is strictly speaking not a limitation of DeepDecipher but rather of the methods themselves, but this points to the broader point that since DeepDecipher does not come with any novel methods, it is limited by the state of current LLM MLP interpretability methods.

The usability of DeepDecipher in practice is currently broadly untested due to time constraints.
This is high on the list of next steps as it is necessary to guide further improvements.

\section{Related works}
\label{sec:appendix-A}

\textbf{Mechanistic interpretability:} \cite{olah_feature_2017} and \cite{olah2020zoom} introduce mechanistic interpretability as the foundational pursuit to reverse-engineer the algorithms learned by neural networks.
When \cite{elhage2021mathematical} defined a mathematical overview of the Transformer models, mechanistic interpretability became more focused on language models \cite{nanda_actually_2023, elhage2022solu, elhage2022toy}.
Existing examples of research work within Transformer mechanistic interpretability includes finding functional circuits for constrained semantic tasks in large language models (LLMs) \cite{wang_interpretability_2022} and taking steps towards automating this process \cite{conmy_towards_2023}, using mechanistic interpretability to define metrics for grokking in LLMs \cite{nanda_progress_2023, power_grokking_2022}, and defining a new softmax linear activation unit \cite{elhage2022solu} to avoid the issues of semantic superposition and polysemanticity present in transformer models \cite{elhage2022toy}.
\cite{foote_neuron_2023} and \cite{bills_language_2023} formulate techniques to build models of an MLP neuron's activation based on the augmented training examples that neuron activates the most for.
\cite{geva_transformer_2021} find that semantic explanations of neuron activity correspond with that neuron's activity, \cite{gurnee_finding_2023} find different groups of neuron activity, and \cite{antverg_pitfalls_2022} critique linear probes and propose new models for analyzing MLP neurons.

\textbf{Research tooling for mechanistic interpretability:} Contemporary interpretability research has two challenges that tools for interpretability can solve; 1) to extract the features and variables we need for research, we often have to rerun models, which can be slow and expensive, and 2) accessibility of large language model internals \cite{schubert_openai_2020}.
\cite{schubert_openai_2020} introduce the OpenAI Microscope, a tool that visualizes the maximum activating examples for layers and neurons in eight vision models.
Inspired by the OpenAI Microscope, \cite{nanda_neuroscope_2022} introduces NeuroScope, a website with access to all MLP neurons' most activating dataset examples for 25 Transformer models.
NeuroScope is built using TransformerLens \cite{nandatransformerlens2022}, a Python package that interfaces with PyTorch \cite{paszke_pytorch_2019} to extract activations on input from Transformer models.
\cite{yeh_attentionviz_2023} introduce a visualization technique for attention that uses query-key embeddings to understand and visualize self-attention mechanisms in Transformers.

\textbf{Model editing:} Mechanistic interpretability seeks to reverse-engineer neural networks to ensure that we understand them before we deploy them in real-world scenarios \cite{olah_feature_2017, amodei2016concrete}.
Techniques exist to update the activation behavior of targeted neurons, e.g., data-based model intervention by fine-tuning \nocite{} or model retrainings \nocite{} along with direct intervention methods, such as ROME \cite{raunak_rank-one_2022}, and MEMIT \cite{meng_mass-editing_2022}.
These methods identify neurons associated with specific semantic associations and perform direct intervention on the activation through ablations \cite{durrani_analyzing_2020}

\section{API usage}\label{sec:api_usage}
Our repository includes a detailed and up-to-date description for how to use DeepDecipher.
However, we provide a short description of the core usage here.
All presented endpoints should be prepended with \verb|https://deepdecipher.org|.

\textbf{Querying:}
Queries use the format
\begin{lstlisting}[style=responsestyle]
/api/<model>/<service>/<layer>/<neuron>
\end{lstlisting}
This returns a JSON object with information specified by \lstinline{service} (e.g. \lstinline{neuroscope} or \lstinline{neuron2graph}) for the specified model, layer, and neuron.
For example
\begin{lstlisting}[style=responsestyle]
r=requests.get('https://deepdecipher.org/api/solu-8l/neuroscope/7/1423')
\end{lstlisting}
returns a JSON object with all the Neuroscope data for the $1423$rd neuron in the $8$th MLP layer of the \lstinline{solu-8l} model.
One then only needs to write
\begin{lstlisting}[style=responsestyle]
snippets = [r.json()["data"]["texts"][k]["tokens"] 
        for k in range(10)]
\end{lstlisting}
to get the tokens of the first $10$ texts for the neuron.
Similarly, querying the layer and model information follows a similar syntax:
\begin{lstlisting}[style=responsestyle]
/api/<model>/<service>/<layer>
/api/<model>
\end{lstlisting}

Here is an example using the neuron store API to \verb|neuron2graph-search| that receives a trimmed and lowercase token and returns lists of neurons that (a) activate the most to that token and (b) whose activation is most affected by this token.
\begin{lstlisting}[style=responsestyle]
/api/solu-6l/neuron2graph-search?query=any:the
\end{lstlisting}
Here, the response will be all neurons that match either (a) or (b) for token \verb|the| since we use the \verb|any| keyword.
In this example, we receive 1976 results.
For comparison \verb|he| returns 254 neurons, \verb|she| return 126 neurons, and \verb|dream| returns 4 neurons in the following format.
\begin{lstlisting}[style=responsestyle]
[
    {layer: 2, neuron: 1917}, 
    {layer: 5, neuron: 2799}, 
    ..., 
    {layer: 5, neuron: 734}
]
\end{lstlisting}

\end{document}